\documentclass[conference]{IEEEtran}
\usepackage{multirow}
\usepackage{pgfplots}
\usepackage{xcolor}
\usepackage{tikz}
\usepackage{graphicx}
\usepackage{caption}
\usepackage{subcaption}

\usetikzlibrary{shadows,arrows.meta,positioning,backgrounds,fit,chains,scopes}

\definecolor{hous}{HTML}{b88b4d}
\definecolor{green}{HTML}{79c561}
\definecolor{farming}{HTML}{ded94c}
\definecolor{trans}{HTML}{b4b4a9}
\definecolor{services}{HTML}{ff362e}
\definecolor{other}{HTML}{dbd4d3}
\definecolor{industry}{HTML}{db79c0}
\definecolor{water}{HTML}{7982db}
\definecolor{techinfra}{HTML}{303355}

\tikzset{%
  materia/.style={draw, fill=blue!20, text width=6.0em, text centered, minimum height=1.5em,drop shadow},
  etape/.style={materia, text width=8em, minimum width=10em, minimum height=3em, rounded corners, drop shadow},
  linepart/.style={draw, thick, color=black!50, -LaTeX, dashed},
  line/.style={draw, thick, color=black!50, -LaTeX},
  ur/.style={draw, text centered, minimum height=0.01em},
  back group/.style={fill=yellow!20,rounded corners, draw=black!50, dashed, inner xsep=15pt, inner ysep=10pt},
}

\newcommand{\transreceptor}[3]{%
  \path [linepart] (#1.east) -- node [above] {\scriptsize #2} (#3);}
\IEEEoverridecommandlockouts
\usepackage{cite}
\usepackage{url}
\usepackage{footnote}
\usepackage{amsmath,amssymb,amsfonts}
\usepackage{algorithmic}
\usepackage{graphicx}
\usepackage{textcomp}
\usepackage{xcolor}
\def\BibTeX{{\rm B\kern-.05em{\sc i\kern-.025em b}\kern-.08em
    T\kern-.1667em\lower.7ex\hbox{E}\kern-.125emX}}

\usepackage{eso-pic}
\newcommand\AtPageUpperMyright[1]{\AtPageUpperLeft{%
 \put(\LenToUnit{0.5\paperwidth},\LenToUnit{-1cm}){%
     \parbox{0.5\textwidth}{\raggedleft\fontsize{9}{11}\selectfont #1}}%
 }}%
\newcommand{\conf}[1]{%
\AddToShipoutPictureBG*{%
\AtPageUpperMyright{#1}
}
}

\IEEEoverridecommandlockouts
\IEEEpubid{
    \hspace{1mm}
    \makebox[\columnwidth]{978-0-7381-2333-2/20/\$31.00 \copyright 2020 IEEE \hfill}
    \hspace{\columnsep}
    \makebox[\columnwidth]{ }
}

\IEEEpubidadjcol

\title{Robust and Consistent Estimation of Word Embedding for Bangla Language by fine-tuning Word2Vec Model}

\begin{document}


\author{
\IEEEauthorblockN{Rifat Rahman}
\IEEEauthorblockA{\textit{Department of Computer Science and Engineering} \\
\textit{Bangladesh University of Engineering \& Technology} and \textit{Southeast University}\\
Dhaka, Bangladesh \\
1405007.rr@ugrad.cse.buet.ac.bd}
}

\maketitle
\conf{2020 23\textsuperscript{rd} International Conference on Computer and Information Technology (ICCIT), 19-21 December, 2020}

\begin{abstract}
Word embedding or vector representation of word holds syntactical and semantic characteristics of a word which can be an informative feature for any machine learning-based models of natural language processing. There are several deep learning-based models for the vectorization of words like word2vec, fasttext, gensim, glove, etc. In this study, we analyze word2vec model for learning word vectors by tuning different hyper-parameters and present the most effective word embedding for Bangla language. For testing the performances of different word embeddings generated by fine-tuning of word2vec model, we perform both intrinsic and extrinsic evaluations. We cluster the word vectors to examine the relational similarity of words for intrinsic evaluation and also use different word embeddings as the feature of news article classifier for extrinsic evaluation. From our experiment, we discover that the word vectors with 300 dimensions, generated from ``skip-gram'' method of word2vec model using the sliding window size of 4, are giving the most robust vector representations for Bangla language.
\end{abstract}


\section{Introduction}
\label{sec:introuction}
Word embedding means the vector representation of every unique word in a multi-dimensional vector space. Word embedding is being used in different sectors of Natural Language Processing (NLP). When we want to build any model, we need to represent each and every data numerically for the model. Because the machine can not understand anything except numeric data. Word embedding gives unique vector representation to every word. 

There exist different deep learning-based models for vectorizing words, such as word2vec, gensim, fasttext, glove, etc. In this study, we use word2vec model, proposed by Mikolov et al. \cite{mikolov2013efficient} for implementing an effective and standard vector representation of words for Bangla language. Word2vec is an unsupervised multi-layer perceptron neural network that can predict a center word from a given set of context words and vice versa.

Several studies for effective or efficient estimation of the dimension or other hyper-parameters of word embedding have been conducted in other languages. Different works in Bangla language have used word embedding as features for different learning-based models, but there is no work related to effective and efficient word embedding. Bangla is a culturally rich language. This is the 7\textsuperscript{th} most spoken language over the world and around 265 million people across the world speak in Bangla as native and non-native speakers\footnote{\url{https://www.dhakatribune.com/world/2020/02/17/bengali-ranked-at-7th-among-100-most-spoken-languages-worldwide}}. Different NLP related tasks, such as document classification, sentiment analysis, named entity recognition, etc need vector representation of texts. For gaining more accuracy in these works in Bangla, we need an effective and efficient estimation of word embedding.

NLP related works in Bangla language are so much challenging. The main reason behind it is lacking effective resources. Again Bangla is a language with rich vocabulary and both grammatical \& syntactical structure of Bangla is complicated. There is no effective built-in pre-processing tool like lemmatizer, parts of speech (POS) tagger, etc for Bangla. Furthermore, there are many words that are morphologically same but are semantically different. However, there are some challenges related to biasness of corpus for word embedding which give poor performance. 

In our literature, we collect data of almost six years from different online news portals. These data include different categories like politics, business, education, literature, etc. We train our word2vec model utilizing these versatile data after pre-processing. We apply both ``Skip-gram'' (SG) and ``Continuous bag of words'' (CBOW) techniques of word2vec model. We learn the word embedding by tuning dimension, window size, learning rate, etc. Then we perform intrinsic and extrinsic experiments utilizing our embeddings. By comparing their performances against the same dataset, we select the effective and standard word embedding.

Several studies in Bangla language use word2vec model as feature extraction technique for classification problems \cite{chowdhury2018comparative, ahmad2016bengali, sumit2018exploring}. Some works present the comparative analysis among different models for word embedding \cite{ritu2018performance}. But there is no study related to the effectiveness and standardization of word embedding.

The rest of the paper is organized as follows. In Section \ref{sec:related_work}, we review different existing literature of word embedding. We describe our corpus and present our proposed method in Section \ref{sec;method}. Section \ref{sec:experimental_result} depicts the experimental results of our proposed method. Finally, we come to a conclusion and describe our future directions and improvements in Section \ref{sec:conclusion}. 

\section{Related Work}
\label{sec:related_work}
There are some works for estimating the dimension of word embedding in English language. Yang et al. \cite{yang2015supervised} have proposed a supervised fine-tuning framework. They have applied supervised fine-tuning after unsupervised learning of word vectors to improve the effectiveness of word embedding.

Das et al. \cite{das2019critical} have experimented to determine the critical dimension of word embedding. They have applied principal component analysis (PCA) \cite{wold1987principal} to reduce the high dimension of word vectors in such a way that the semantic and syntactic meanings with the reduced dimension would be the same as the meanings of original high dimensional word embedding. By eliminating the common mean vector and a few top principal components from the word vectors, Mu et al. \cite{mu2017all} have implemented a simple post-processing algorithm to represent strong and effective word embedding. Raunak et al. \cite{raunak2019effective} combined the previous post-processing algorithms with PCA-based dimension reduction to build effective word embedding. 

The work of Bolukbasi et al. \cite{bolukbasi2016man} proposed a debiasing method to reduce the biasness from word embedding. Shu et al. \cite{shu2017compressing} have presented a deep compositional discrete code learning model to compress word embedding. For strengthening salient information and weakening noise in the original word embeddings, Nguyen et al. \cite{nguyen2016neural} proposed a deep feed-forward neural network filter.

Several studies have performed comparative analysis among different word embedding models. Wang et al. \cite{wang2019evaluating} discussed six different models for word embedding. They performed intrinsic and extrinsic evaluation methods and found that different evaluators gave the best performance depending on different aspects of word embedding models. According to the investigation among three word embedding models done by Naili et al. \cite{naili2017comparative} for topic segmentation in both English and Arabic language, word2vec provides the best performance. Suleiman et al. \cite{suleiman2018comparative} also performed comparative studies among different word embedding methods.

There is no study related to standardization of the dimension or other parameters for word embedding in Bangla language. Most of the works only use word embedding models for feature extraction, but they have not shown standard estimation of word embedding. There are few comparative studies among different models for Bangla language. Chowdhury et al. \cite{chowdhury2018comparative} have performed comparative analysis of different word embedding models by a specific extrinsic classification problem. Ritu et al. \cite{ritu2018performance} have also discussed the comparative performance analysis of several word embedding models by intrinsic evaluation.

\section{Methodology}
\label{sec;method}
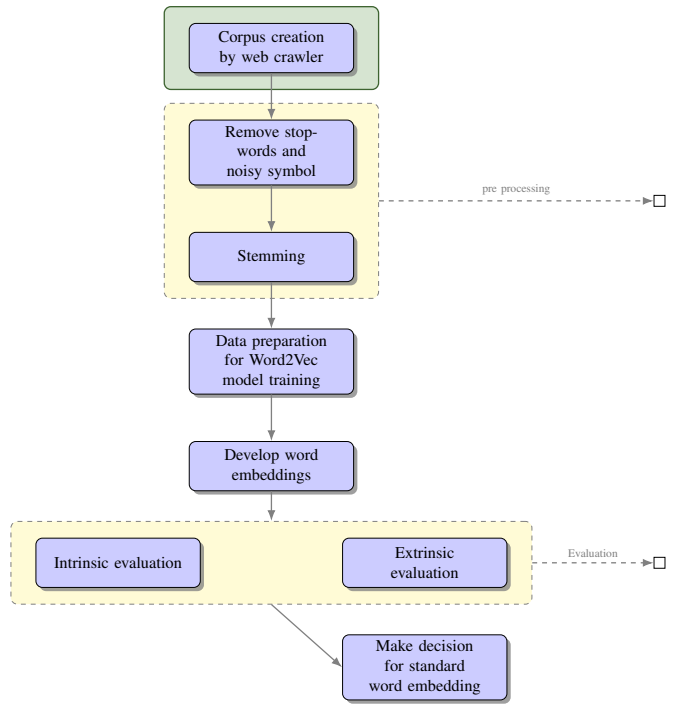
\begin{figure}
\centering
\resizebox{\columnwidth}{!}{
\begin{tikzpicture}
  [
    start chain=p going below,
    every on chain/.append style={etape},
    every join/.append style={line},
    node distance=1 and -.25,
  ]
  {
    \node [on chain, join] {Corpus creation by web crawler};
    \node [on chain, join] {Remove stop-words and noisy symbol};
    \node [on chain, join] {Stemming};
    \node [on chain, join] {Data preparation for Word2Vec model training};
    \node [on chain, join] {Develop word embeddings};
    {[start branch=r going below right]
      \node [on chain] {Extrinsic evaluation};
    }
    {[start branch=l going below left]
      \node [on chain] {Intrinsic evaluation};
    }
    {[continue branch=r going below]
      \node [on chain] {Make decision for standard word embedding};
    }
  }

  \begin{scope}[on background layer]
    \node (bk1) [back group] [fit=(p-2) (p-3)] {};
    \node (bk3) [back group] [fit=(p/r-2) (p/l-2)] {};
    \node [draw, thick, green!50!black, fill=green!75!black!25, rounded corners, fit=(p-1), inner xsep=15pt, inner ysep=10pt] {};
  \end{scope}

  \path [line] (p-5.south) --  (bk3.north);
  \path [line] (bk3.south) --  (p/r-3.west);

  \path (bk1.east)+(+6.0,0) node (ur1)[ur] {};
  \node (ur3)[ur] at (bk3.east -| ur1) {};
  
  \transreceptor{bk1}{pre processing}{ur1};
  \transreceptor{bk3}{Evaluation}{ur3};
\end{tikzpicture}
}
\caption{Work flow of the proposed method}
\label{fig:proposed_method}
\end{figure}

In this section, we provide a brief description of our proposed methods. We create our corpus from different online news portals and Wikipedia with different contexts. Then we perform pre-processing techniques and build our word2vec model. We apply both ``skip-gram'' (SG) and ``continuous bag of words'' (CBOW) techniques for our model. Then we conduct both intrinsic and extrinsic evaluations. Figure \ref{fig:proposed_method} is presenting the workflow diagram of our proposed method.

\subsection{Corpus Creation}
\label{sub:corpus}
We implement web crawling program to fetch raw data from the HTML pages. We use urllib package and BeautifulSoup library of python language for this purpose. We collect data from several popular online news portals of Bangladesh like Prothom Alo, Kaler Kantho, etc, and Wikipedia.
\begin{table}[ht]
\caption{Category distribution of our Corpus}
\label{tab:category_dist}
\begin{tabular}{||c|c|||c|c||}
\hline
\hline
\textbf{Category}&\textbf{Number of articles}&\textbf{Category}&\textbf{Number of articles}\\
\hline
\hline
Politics&25,372&Entertainment&24,377\\
\hline
Crime&17,522&Bio-graphy&14,178\\
\hline
Accident&22,969&Religion&13,363\\
\hline
Education&15,908&Technology&23,785\\
\hline
Economy&24,567&sports&23,146\\
\hline
Literature&18,335&Life-style&20,219\\
\hline
\hline
\end{tabular}
\end{table}
 We collected data from different contexts including politics, literature, technology, economy, etc to reduce the bias and to estimate standard word embedding for Bangla language. Table \ref{tab:category_dist} is presenting the category distribution of our corpus. We collect data from almost six years dated from 2015 to early 2020.

Our corpus has a huge collection of articles. The corpus includes 243,741 articles with 4,776,708 sentences from different field of context. We have shown the statistical overview of our corpus in Table \ref{tab:stat_corpus}.

\begin{table}[h]
\caption{Statistical overview of our corpus}
\label{tab:stat_corpus}
\begin{center}
\begin{tabular}{||c||c||}
\hline
\hline
\textbf{Parameters}&\textbf{Total Number}\\
\hline
\hline
Articles&243,741\\
\hline
Sentences&4,776,708\\
\hline
Average Sentences per Article&19.6\\
\hline
Words&54,167,865\\
\hline
Unique Words (Vocabulary size)&726,430\\
\hline
Average words per sentence&11.34\\
\hline
Average words per article&223.26\\
\hline
\hline
\end{tabular}
\end{center}
\end{table}

\subsection{Pre-processing}
\label{sub:preprocess}
After creating the corpus, we apply different pre-processing methods because the corpus contains unstructured data. Again there are many noises like characters from other languages, different symbols, URL links, etc. 

The pre-processing techniques include both sentence \& word tokenizing. We clean our data from noisy symbols and characters. Then we eliminate stop-words and connecting words. There are several resources of Bangla stop-words\footnote{\url{https://github.com/stopwords-iso/stopwords-bn}} and connecting words. Finally, we apply stemming to get the root words. We use Bangla stemmer\footnote{\url{https://pypi.org/project/bangla-stemmer/}} function of python language. Stemming shrinks the size of total vocabulary because semantically same words can be in form of different representations.

Finally, after the pre-processing, the corpus includes separate sentences which are containing clean words. We use these lists of sentences for training our word2vec model.

\subsection{Data Preparation}
\label{sub:data_prep}
After pre-processing our corpus, we need to prepare data to feed the model because machines can not understand natural language except numeric values. For data preparation, we represent each word as one hot encoding vector at first. From Figure \ref{fig:w2v}, the dimension of the one hot encoding vector is $V \times 1$. $V$ refers to vocabulary size and for our study, the value of $V$ is 726,430. We tune the window size from 2 to 4. Based on the window size, we take the one hot encoding vector of center word and corresponding average one hot encoding vector of the context words. Thus, we prepare data by depending on different window sizes from the processed corpus.

\subsection{Model Architecture}
\label{sub:w2v}
Word2vec is a unsupervised shallow neural network based model which was developed by Mikolov et al. \cite{mikolov2013efficient}. at Google.

\begin{figure}[h]
\centerline{\includegraphics[width=\linewidth]{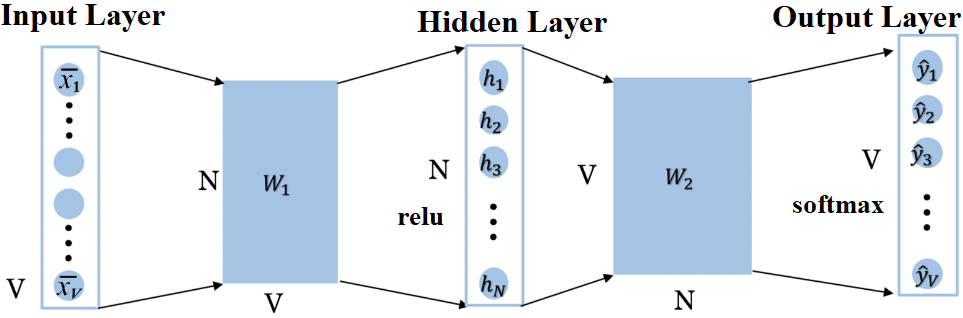}}
\caption{Architecture of Word2Vec model}
\label{fig:w2v}
\end{figure}
Figure \ref{fig:w2v} depicts the architecture of Word2Vec model. This model is a 2-layer neural network. The input of the model is one hot encoding vector of center word or the average one hot encoding vector of context words with dimension $V \times 1$. The number of neurons of the hidden layer is the same as dimension size, $N$. We tune the size of the dimension valued from 100 to 500 with an interval of 100. The number of hidden nodes for the output layer is equal to the vocabulary size, $V$. All three layers are fully connected. The activation function, used in the hidden layer, is $relu$ and the activation function for the output layer is $softmax$. The dimension of the first weight matrix, $W_{1}$ is $N \times V$ and the second weight matrix, $W_{2}$ is $V \times N$. We transpose the first weight matrix and take average with the second one. Finally, we get an average weight matrix that has dimension $V \times N$. Thus we learn different word embeddings by tuning dimension and sliding window size. So, from Figure \ref{fig:w2v}, we are getting,
\begin{displaymath}
h = ReLU(W_{1}X+b_{1})
\end{displaymath}
\begin{displaymath}
\hat y = softmax(W_{2}h+b_{2})
\end{displaymath}
Here, $W_{1}$ \& $W_{2}$ are weight matrices, $b_{1}$ \& $b_{2}$ are bias matrices,  and $X$ is input.

We feed the prepared data from Subsection \ref{sub:data_prep} to the word2vec model. Depending on the input, word2vec has two implementation methods.

\paragraph{Continuous Bag of Words (CBOW)} ``CBOW'' takes context word as input and predicts single center word. So, in this implementation, the average one hot encoding vector of the context words is feed to input layer for the prediction of center word.

\paragraph{Skip-gram (SG)} ``SG'' takes center word as input and then predict the probabilities of the context words. So, this implementation takes one hot encoding vector of center word as input for the prediction of context words.

For reducing the overfitting, we use ``adam'' optimizer with learning rate as 0.01 and we evaluate ``categorical cross-entropy loss'' for cost computation. We take the batch size as 256.

\subsection{Intrinsic Evaluation}
\label{sub:intrinsic}
We use a Bangla wordnet\footnote{\url{https://github.com/soumenganguly/Bangla-Wordnet}} as the reference for the semantic performance analysis of all the word embeddings. We fetch the synonyms set of some most frequent nouns and verbs of our corpus from the wordnet. We conduct this experiment on a total of 50 words with 5 classes. Then we compare the ground truth classes with the clusters generated from different word embeddings. 

Before clustering and visualization, we apply ``Principal Component Analysis'' (PCA) \cite{wold1987principal} to reduce the dimension to 2. Then we cluster all the words using k-means clustering algorithm \cite{jain2010data} to visualize the semantic relationship of words. Thus we perform the intrinsic evaluation.

\subsection{Extrinsic Evaluation}
\label{sub:extrinsic}
For extrinsic evaluation, we implement a neural network-based news article classifier. We use our different implemented word embeddings as the feature of the classifier. We build a labeled dataset from our corpus. In the dataset, we map articles with five categories including state, international, economy, sports, and entertainment. We take total 100,000 news articles in the dataset.

We set our article length as 400 words. We pad with zero for shorter lengthen articles. Then we create an embedding layer. We utilize our different pre-trained word embeddings induced from Subsection \ref{sub:w2v} for different experiments. After that, we add 1D convolutional layer \cite{kalchbrenner2014convolutional} with 128 filters with $relu$ activation function. We take the kernel size as 3 and add global max pooling layer with pool size 2. We set the dropout rate as 0.8 and add a flatten layer. Then we add a hidden dense layer with 64 nodes \& $relu$ activation function and finally add output layer with 5 nodes \& $softmax$ activation function.

We set ``adam'' optimizer with learning rate 0.01 and ``sparse categorical cross-entropy'' loss function for the training model. We set the batch size as 256 and total epochs as 10. We split 10\% of our dataset for testing and among the rest of the dataset, 90\% of them is used as the training set and 10\% of them is used as validation dataset.

\section{Experimental Result}
\label{sec:experimental_result}
In this section, we discuss the comparative performance analysis of different word embeddings induced by two techniques of Word2Vec model with different dimensions and window sizes. Then we present the standard word embedding by analyzing the results of both intrinsic and extrinsic evaluations. 

\subsection{Experimental Setup}
\label{experiment}
For our experiment, we use Python programming language on ``Google Colab''\footnote{\url{https://colab.research.google.com/notebooks/intro.ipynb}} platform. ``Google Colab'' provides both GPU and TPU services. For data representation and visualization, we use python libraries like numpy, pandas, matplotlib etc. For clustering and dimensionality reduction, we take help from Scikit-learn\footnote{\url{https://scikit-learn.org/stable/}} library. We also utilize  Keras library with Tensorflow background for the extrinsic evaluation.

\subsection{Performance Metrics}
\label{perform_met}
\paragraph{Intrinsic evaluation} For intrinsic evaluation, we calculate purity, normalized mutual information (NMI) of k-means cluster using different word embeddings. Suppose, $\Omega=\{ \omega_1, \omega_2, \ldots, \omega_K \}$ is the set of clusters and $\mathbb{C}=\{ c_1,c_2,\ldots,c_J \}$ is the set of classes of true labels for $N$ samples, then the formulas\footnote{\url{https://nlp.stanford.edu}} are:
\begin{displaymath}
\mbox{purity}(
\Omega,\mathbb{C}
) =
\frac{1}{N}
\sum_k \max_j
\vert\omega_k \cap
c_j\vert
\end{displaymath}
\begin{displaymath}
\mbox{NMI}(\Omega , \mathbb{C})
=
\frac{
I(\Omega ; \mathbb{C})
}
{
[H(\Omega)+ H(\mathbb{C} )]/2
}
\end{displaymath}
Here, $I$ refers to mutual information and $H$ refers to entropy.
\paragraph{Extrinsic evaluation} We evaluate Precision, Recall and F1-score to measure the performances of different word embeddings applying as feature to neural network based model. The formulas of these performance metrics are:
\begin{displaymath}
Precision=\frac{True~Positive}{True~Positive~+~False~Positive}
\end{displaymath}
\begin{displaymath}
Recall=\frac{True~Positive}{True~Positive~+~False~Negative}
\end{displaymath}
\begin{displaymath}
F1{\text-}score=\frac{2\times(Recall\times Precision)}{Recall+Precision}
\end{displaymath}

\subsection{Result Analysis}
\label{sub:result_analysis}
We can observe the results of all experiments for both intrinsic and extrinsic evaluation from Table \ref{tab:performance}. Total 15 rows of Table \ref{tab:performance} represent experiments with different window sizes and dimensions of different word embeddings. 

\begin{table*}[h]
\caption{Performance Measure of different word embeddings methods of Word2Vec for both intrinsic \& extrinsic evaluation}
\label{tab:performance}
\begin{center}
\begin{tabular}{|c|c||c|c||c|c|c||c|c||c|c|c|}
\hline
\multirow{2}{*}{\textbf{Window}}&\multirow{3}{*}{\textbf{Dimension}}&\multicolumn{5}{|c||}{\textbf{CBOW}}&\multicolumn{5}{|c|}{\textbf{SG}}\\
\cline{3-12}
\multirow{2}{*}{\textbf{size}}&&\multicolumn{2}{|c||}{\textbf{Intrinsic Evaluation}}&\multicolumn{3}{|c||}{\textbf{Extrinsic Evaluation}}&\multicolumn{2}{|c||}{\textbf{Intrinsic Evaluation}}&\multicolumn{3}{|c|}{\textbf{Extrinsic Evaluation}}\\
\cline{3-12}
&&\textbf{Purity}&\textbf{NMI}&\textbf{Precision}&\textbf{Recall}&\textbf{F1-score}&\textbf{Purity}&\textbf{NMI}&\textbf{Precision}&\textbf{Recall}&\textbf{F1-score}\\
\hline
\hline
\multirow{5}{*}{2}&100   &0.66&0.563  &0.885&0.842&0.863  &0.68&0.587 &0.911&0.886&0.898\\
\cline{2-12}
&200  &0.7&0.602  &0.861&0.856&0.859   &0.71&0.607 &0.845&0.903&0.871\\
\cline{2-12}
&300  &0.75&0.631   &0.933&0.905&0.917   &0.78&0.659 &0.954&0.909&0.93\\
\cline{2-12}
&400  &0.68&0.576  &0.921&0.902&0.911   &0.7&0.601 &0.956&0.928&0.947\\
\cline{2-12}
&500  &0.71&0.589   &0.94&0.901&0.919    &0.74&0.625 &0.984&0.905&0.94\\
\hline
\hline
\multirow{5}{*}{3}&100   &0.69&0.578  &0.919&0.869&0.892  &0.73&0.612   &0.959&0.941&0.95\\
\cline{2-12}
&200   &0.67&0.571   &0.854&0.852&0.853     &0.7&0.605 &0.863&0.859&0.861\\
\cline{2-12}
&300   &0.78&0.639  &0.925&0.874&0.897     &0.81&0.667 &0.943&0.878&0.907\\
\cline{2-12}
&400   &0.77&0.631   &0.873&0.869&0.871    &0.82&0.667 &0.887&0.884&0.886\\
\cline{2-12}
&500    &0.73&0.607  &0.831&0.827&0.829   &0.75&0.629 &0.878&0.84&0.858\\
\hline
\hline
\multirow{5}{*}{4}&100  &0.81&0.668 &0.907&0.904&0.905   &0.84&0.695   &0.948&0.915&0.931\\
\cline{2-12}
&200  &0.86&0.739  &0.922&0.899&0.91   &0.88&0.761 &0.921&0.923&0.922\\
\cline{2-12}
&300  &0.87&0.762  &0.935&0.927&0.931   &\textbf{0.9}&\textbf{0.799} &\textbf{0.999}&\textbf{0.935}&\textbf{0.964}\\
\cline{2-12}
&400  &0.81&0.67  &0.93&0.9&0.914   &0.86&0.736 &0.964&0.917&0.931\\
\cline{2-12}
&500  &0.74&0.627  &0.907&0.905&0.906    &0.76&0.653 &0.934&0.907&0.92\\
\hline
\end{tabular}
\end{center}
\end{table*}

\subsubsection{Comparison between ``SG'' and ``CBOW'' methods}
\label{subsub:SG_CBOW}
\begin{figure}[h]
    \centering
    \minipage{0.85\linewidth}
      \includegraphics[width=\linewidth]{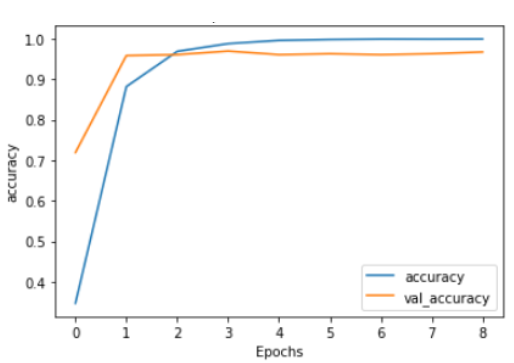}
      \subcaption{`SG' method [Dimension: 300, Window size: 4]}
      \label{fig:sg_accuracy}
    \endminipage\hfill
    \minipage{0.85\linewidth}
      \includegraphics[width=\linewidth]{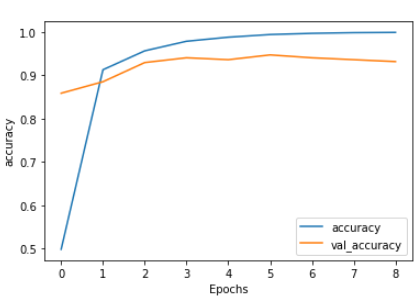}
      \subcaption{`CBOW' method [Dimension: 300, Window size: 4]}
      \label{fig:cbow_accuraacy}
    \endminipage
    \caption{Epochs vs training \& validation accuracy for news article classification for both methods of word2vec}
    \label{fig:SG_CBOW_epochs_vs_accuracy}
\end{figure}
From Table \ref{tab:performance}, we find that ``Skip-gram'' method is giving better performance than ``continuous bag of words'' method for both intrinsic and extrinsic experiments. This is because ``skip-gram'' method performs well when corpus is not large enough. Again from Figure \ref{fig:SG_CBOW_epochs_vs_accuracy} we can discover that the validation accuracy of ``SG'' method is fine while the same curve is moving downwards with the increment of epochs. We get these findings from the experiment in which we use the dimension and window size as 300 and 4 respectively that combination gives the best performance among all other word embeddings.


\subsubsection{Window size dependent performance}
\label{subsub:win_performance}

\begin{figure}[h]
    \centering
    \begin{tikzpicture}
        \begin{axis}[
            ymin=0,ymax=1.08,
            xtick=data,
            enlarge x limits={abs=2cm},
            height = 6cm,
            width =8.5cm,
            major x tick style = transparent,
            ybar=2*\pgflinewidth,
            bar width=19pt,
            ymajorgrids = true,
            symbolic x coords={Average score,Maximum score},
            ylabel= F1-score,
            ytick align=outside,
            ytick pos=left,
            major x tick style = transparent,
            legend style={
                at={(1,1.1)},
                anchor=south east,
                column sep=1ex
            },
            nodes near coords,
            nodes near coords align={vertical}
            ]
            \addplot[ybar,fill=green] coordinates {
                (Average score,.91) (Maximum score,0.92)};
            \addplot[ybar,fill=red] coordinates {
                (Average score,.89) (Maximum score,0.95)};
            \addplot[ybar,fill=blue] coordinates {
                (Average score,.94) (Maximum score,0.96)};
         \legend{Window size: 2, Window size: 3, Window size: 4}
        
        \end{axis}
    \end{tikzpicture}
    \caption{Average and maximum F1-score with different window sizes for `SG' method}
    \label{graph:window_f1_score}
\end{figure}
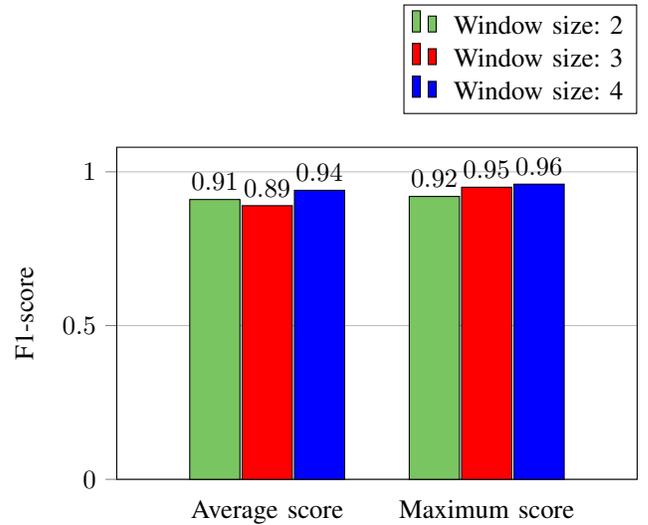
Figure \ref{graph:window_f1_score} depicts the average and maximum F1-score of five dimensions for three different window sizes for `SG' method. Word embedding, which is induced from window size 4, is giving an excellent F1-score for news article classification. Again Table \ref{tab:performance} presents that the window size of 4 is giving the best performance for both intrinsic and extrinsic evaluations. 

\subsubsection{Performance among different dimensions}
\label{subsub:dim_performance}
\begin{figure}[h]
    \centering
    \includegraphics[width=\linewidth]{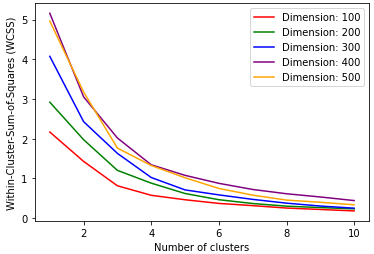}
    \caption{Elbow method of clustering using different dimensions of word vector [`SG' method, Window size: 4]}
    \label{fig:elbow}
\end{figure}
We apply elbow method for the curves of different dimensions with ``SG'' technique and window size 4. Figure \ref{fig:elbow} presents the graph of ``number of clusters'' vs ``within cluster sum of squares'' (WCSS). WCSS refers to the summation of the squares of the distances between a point and it's cluster based centroid. From all the five curves of Figure \ref{fig:elbow}, the elbow of the curve of 300 dimensional word embedding marks 5 clusters which is consistent with our assumption of 5 true classes for intrinsic evaluation. We also find the robustness of 300 dimensional word vectors from Table \ref{tab:performance}.

\subsection{Result Summary}
\begin{figure}[h]
    \centering
    \includegraphics[width=\linewidth]{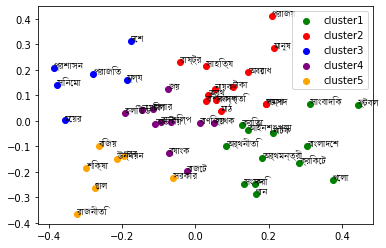}
    \caption{Word visualization of 50 words from our corpus with the standard word embedding}
    \label{fig:word_visualization}
\end{figure}

\begin{figure}[h]
    \centering
    \includegraphics[width=\linewidth]{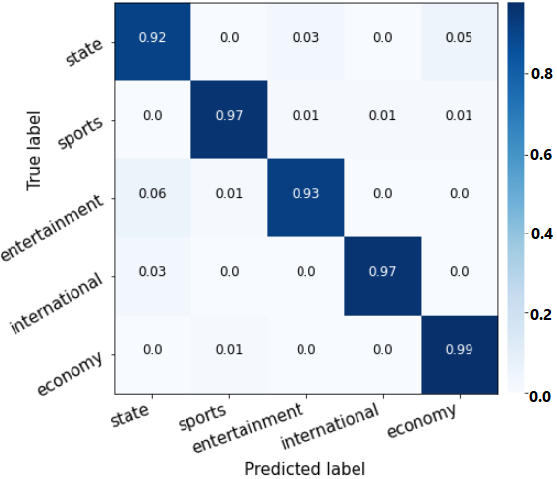}
    \caption{Confusion matrix of news article classification}
    \label{fig:confusion_matrix}
\end{figure}

By applying several experiments on different word embeddings, we discover that ``skip-gram'' technique of word2vec with 300 dimensions and 4-gram is giving the effective output for both intrinsic and extrinsic evaluations. Figure \ref{fig:word_visualization} is showing the consistency of the K-Means clusters with ground truth classes. We get 90\% purity and 79.9\% NMI from these clusters. We can also observe the confusion matrix for news article classification from Figure \ref{fig:confusion_matrix}. The classifier, utilizing the most effective word embedding is giving 96.4\% F1-score.
\label{sub:summary}

\section{Conclusion}
\label{sec:conclusion}
Word embedding is an informative feature for machine learning methods that carries semantic and syntactic information of words. We implement both methods of Word2Vec model for Bangla language and tune the dimension and window size. We find that word embedding from skip-gram (SG) method with the dimension of 300 \& window size of 4 is giving the best performance. We have gotten 90\% purity and 99.9\% precision from intrinsic and extrinsic evaluation respectively. These vector representations of Bangla words can be used in any NLP related tasks as a standard word embedding. In the future, we will enlarge our corpus for better word embedding. Again, we have a plan to perform the same experiments on other word embedding models like fasttext, glove, gensim, etc, and do a comparative study among all the standard word embeddings from different models for Bangla. Furthermore, we have a scheme to investigate critical and efficient vector representations of words in the future.


\end{document}